\documentclass[letterpaper]{article} 
\usepackage{aaai2026}  
\usepackage{times}  
\usepackage{helvet}  
\usepackage{courier}  
\usepackage[hyphens]{url}  
\usepackage{graphicx} 
\usepackage{natbib}  
\usepackage{caption} 
\usepackage{algorithm}
\usepackage{algorithmic}
\usepackage{subcaption}
\usepackage{amssymb}
\usepackage{bm}
\usepackage{amsmath}
\usepackage{booktabs}
\usepackage{xcolor}
\usepackage{utfsym}
\usepackage{textcomp}

\usepackage{newfloat}
\usepackage{listings}
\DeclareCaptionStyle{ruled}{labelfont=normalfont,labelsep=colon,strut=off} 
\floatstyle{ruled}
\newfloat{listing}{tb}{lst}{}
\floatname{listing}{Listing}
\nocopyright

\title{Segmenting and Understanding: Region-aware Semantic Attention for Fine-grained Image Quality Assessment with Large Language Models}
\author{
    Chenyue Song, Chen Hui, Haiqi Zhu, Feng Jiang, Yachun Mi, Wei Zhang, Shaohui Liu
}
\affiliations{
    Harbin Institute of Technology\\

    Harbin, China\\
    
}

\usepackage{bibentry}

\begin{document}

\maketitle

\begin{abstract}
No-reference image quality assessment (NR-IQA) aims to simulate the process of perceiving image quality aligned with subjective human perception.  However, existing NR-IQA methods either focus on global representations that leads to limited insights into the semantically salient regions or employ a uniform weighting for region features that weakens the sensitivity to local quality variations. In this paper, we propose a fine-grained image quality assessment model, named RSFIQA, which integrates region-level distortion information to perceive multi-dimensional quality discrepancies.  To enhance regional quality awareness, we first utilize the Segment Anything Model (SAM) to dynamically partition the input image into non-overlapping semantic regions. For each region, we teach a powerful Multi-modal Large Language Model (MLLM) to extract descriptive content and perceive multi-dimensional distortions, enabling a comprehensive understanding of both local semantics and quality degradations. To effectively leverage this information, we introduce Region-Aware Semantic Attention (RSA) mechanism, which generates a global attention map by aggregating fine-grained representations from local regions.  In addition, RSFIQA is backbone-agnostic and can be seamlessly integrated into various deep neural network architectures. Extensive experiments demonstrate the robustness and effectiveness of the proposed method, which achieves competitive quality prediction performance across multiple benchmark datasets.
\end{abstract}

\section{Introduction}
\label{intro}
No-reference image quality assessment (NR-IQA) aims to estimate perceptual image quality in a manner consistent with the human visual system (HVS) \cite{ke2021musiq}. Such evaluation can effectively improve the alignment with human perception in various applications, including image compression \cite{cai2024phocolens}, restoration \cite{chen2024prompt}, editing \cite{fang2020perceptual}, and generation \cite{li2023agiqa}. With the emergence of Multi-modal Large Language Models (MLLMs) \cite{bai2025qwen2, liu2023visual, zhu2023minigpt}, MLLM-based IQA methods have gained increasing attention \cite{chen2024q, wu2024q, wu2024towards}. These approaches leverage MLLMs to describe image quality through natural language, recognizing that language more effectively captures human subjective perception and the inherent complexity of IQA tasks \cite{you2024depicting}.
\begin{figure}[t]
\centering
\includegraphics[width=0.43\textwidth]{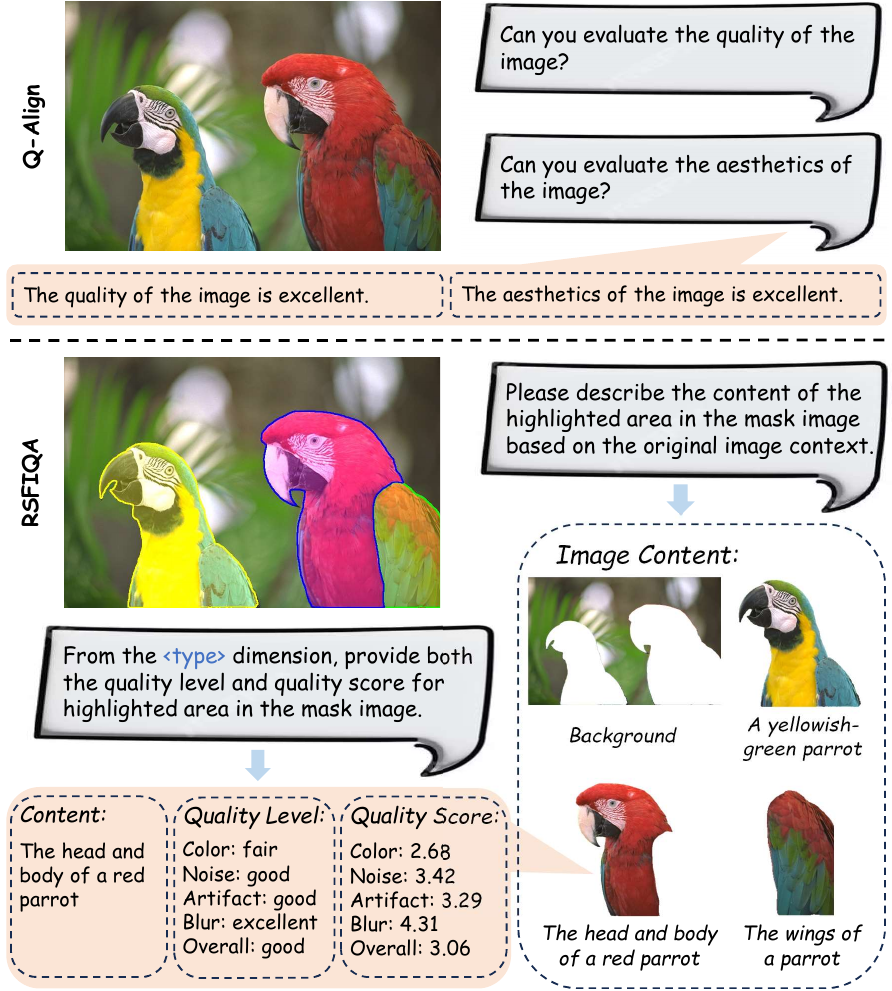}
\caption{Illustration of Region-aware Semantic Guidance in our RSFIQA. It is noticeable that the parrots are clear, while the background is blurry. Prior work such as \cite{wu2023q} conducts an overall image quality assessment, overlooking distortion variations across different semantic regions. In contrast, we segment the image into distinct semantic regions and perform region-wise analysis of content along with multi-dimensional distortion ratings and scores.}
\label{inno}
\end{figure}

However, for quality assessment, a precise numerical score is required to quantify image quality.  Existing description-centric MLLMs underperform traditional IQA methods in accuracy \cite{wu2024q, wu2024towards, you2025teaching}, limiting their utility in real-world IQA applications. Given the strong generalization capability of MLLMs in various downstream tasks and their robust performance in quality description, MLLMs hold potential for image quality scoring. Therefore, in this work, we aim to leverage the inherent adaptability of MLLMs to derive accurate quality descriptions, thereby facilitating the regression of quality scores.

Our investigation reveals that a primary challenge in employing MLLMs for precise quality score prediction lies in their global descriptions potentially overlooking localized distortion characteristics. Previous approaches, like Q-Align \cite{wu2023q}, employed direct question-answering mechanisms to generate global quality descriptions. However, such holistic descriptions incur information loss, resulting in imprecise quality assessment across distinct semantic regions. For example, as illustrated in Figure \ref{inno}, the two parrots are characterized by high sharpness and vivid coloration, whereas the background appears noticeably blurred and underexposed. However, Q-align provides only two global quality scores, inevitably overlooking the quality variations across different regions, which compromises the overall assessment accuracy.

To address the quality variations among semantic regions, we propose a region-level semantics-guided fine-grained IQA approach.  As illustrated in Figure \ref{inno}, rather than employing the MLLM to produce a global quality score, we first utilize the SAM model for automated image semantic segmentation, followed by independent multi-dimensional distortion assessment for each semantic region conducted via the MLLM.  The region-level decomposition enables more precise characterization of local quality variations, particularly effective for cases with non-uniform distortion distributions across semantic regions.  A core innovation of our work is the Region-aware Semantic Attention (RSA) mechanism, which conducts attention computation exclusively within the segmented semantic regions.  Due to its semantic-guided selective attention mechanism, RSA is interpretable and ensures that only pixels within the same semantic region complement each other, thereby eliminating interference from irrelevant regions. We conduct comprehensive experimental comparisons on IQA datasets, and our RSFIQA demonstrates better or competitive performance compared to state-of-the-art methods. Our contributions can be summarized as follows:
\begin{itemize}
\item We propose a region-level semantics-guided fine-grained IQA framework that uses MLLM to generate local multi-dimensional quality descriptions for each semantic region, thereby achieving detailed distortion awareness beyond global ratings.
\item We propose a novel Region-aware Semantic Attention (RSA) mechanism that computes self-attention exclusively within SAM-segmented semantic regions, ensuring the attention operation is confined to the same semantic region and effectively excluding interference from irrelevant regions.
\item The proposed method exhibits strong generality, as its fine-grained perceptual analysis of different semantic regions does not depend on the backbone network structure.  This property allows the method to be applied to various deep backbone networks, thereby improving its adaptability and effectiveness across different image quality assessment scenarios.
\item Extensive experiments on widely used IQA datasets demonstrate that the proposed RSFIQA model is capable of performing region-level fine-grained image quality assessment and achieves state-of-the-art performance.
\end{itemize}
\section{Related Work}
\subsection{No-Reference Image Quality Assessment}
Unlike the full-reference IQA methods, NR-IQA methods can only use low-quality (LQ) images as input to measure image quality without any reference directly. Initially, handcrafted features based on natural image statistics were adopted \cite{mittal2012no, mittal2012making, moorthy2011blind}. Subsequently, deep learning-based approaches \cite{zheng2021learning, shin2024blind, li2024blind, shi2025visual, saha2023re} replaced handcrafted features by learning quality priors directly from human-annotated IQA datasets. Recent studies have further enhanced the accuracy of quality regression by introducing multi-dataset co-training strategy \cite{zhang2021uncertainty}, multi-scale features \cite{ke2021musiq, xu2024boosting}, CLIP pre-training \cite{wang2023exploring}, multidimension attention \cite{yang2022maniqa}, multitask learning \cite{zhang2023blind}, and so on.
\subsection{Multi-modal Large Language Models}
MLLMs integrate the visual modality into large language models (LLMs) \cite{achiam2023gpt, chiang2023vicuna, team2023internlm}, enabling general visual understanding by leveraging the broad training and strong generalization capabilities of LLMs. Existing MLLMs \cite{liu2023visual, hurst2024gpt, zhu2023minigpt} have demonstrated general visual capabilities and can handle various multimodal tasks, such as visual question answering \cite{goyal2017making, liu2024mmbench, lu2022learn}, image captioning \cite{agrawal2019nocaps, chen2015microsoft, young2014image}, document understanding \cite{masry2022chartqa, mathew2021docvqa, singh2019towards}, etc. Given the advanced performance of MLLMs in these high-level perceptual tasks, it is natural to explore the performance of MLLMs in low-level perceptual tasks that are highly related to IQA.
\begin{figure*}[t]
\centering
\includegraphics[width=1.9\columnwidth]{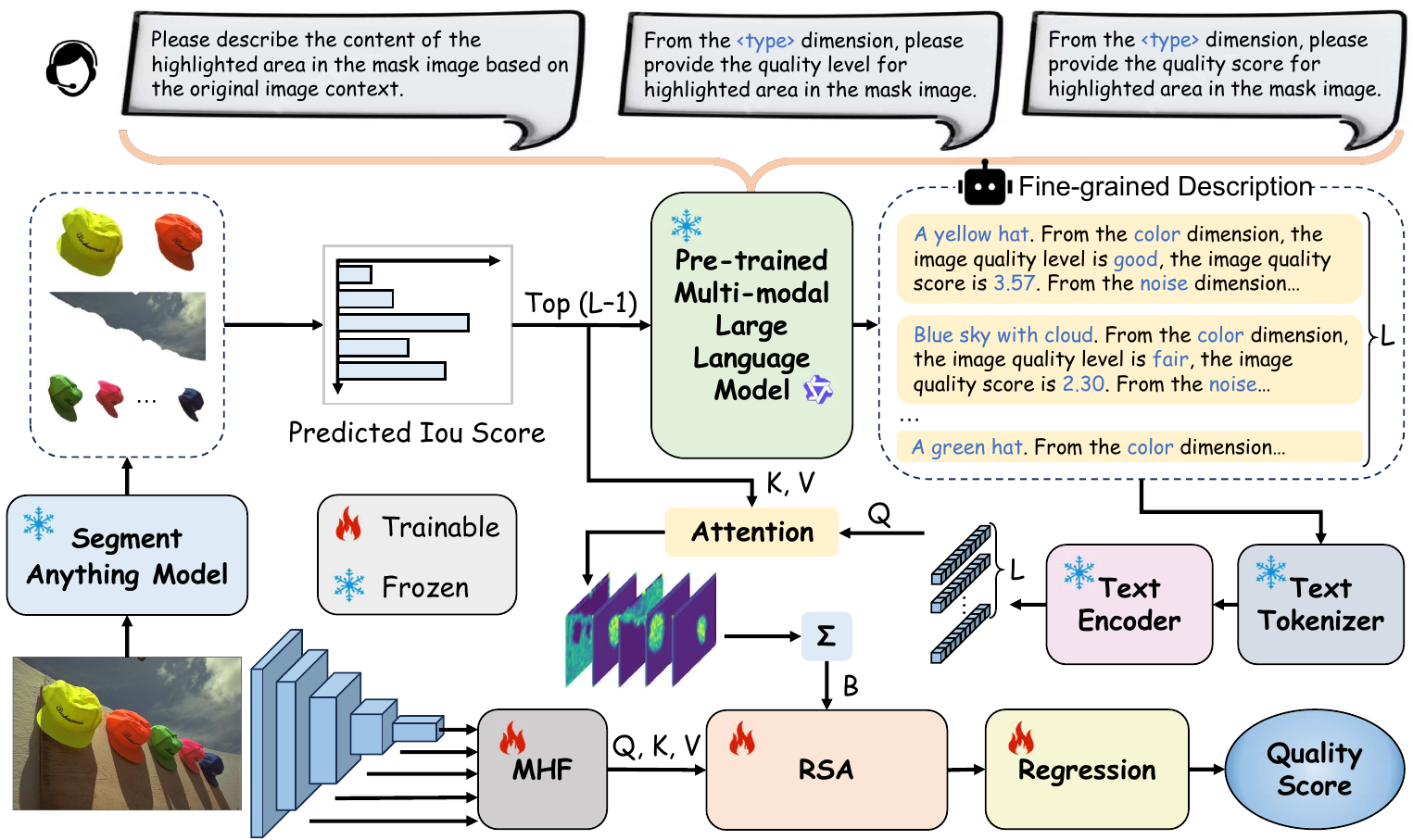} 
\caption{An overview of proposed RSFIQA model. Our model adopts a dual-branch architecture, consisting of a multi-scale image feature extraction branch and a fine-grained distortion perception branch.  Region-level semantic guidance is integrated into the image features via Region-Aware Semantic Attention (RSA).}
\label{fig1}
\end{figure*}
\subsection{MLLM-based IQA methods}
MLLM-based IQA methods leverage the prior knowledge encoded in MLLMs to achieve improved performance in image quality assessment or to produce more fine-grained evaluation results \cite{wu2024comprehensive, zhang2024quality}. Q-Bench \cite{wu2023q} introduces a binary softmax strategy that allows MLLMs to produce an overall image quality score by predicting between two discrete quality levels (i.e., good or poor). This strategy is adopted by Q-Instruct \cite{wu2024q} and Co-Instruct \cite{wu2024towards}. Q-Align \cite{wu2024q} employs one-hot encoded labels to discretize quality scores into five levels for training MLLMs, thereby enabling more accurate quality score prediction. However, as discussed in the Introduction, directly assessing the entire image tends to overlook quality variations across different semantic regions. These models exhibit limited capability in perceiving regional quality and also struggle to support multi-dimensional fine-grained quality evaluation.
\section{Methodology}
In this section, we provide a detailed introduction to the proposed RSFIQA, which uses region-level semantic-guided multi-dimensional distortion information to assist in predicting quality scores.
\subsection{Overall Architecture}
As illustrated in Figure \ref{fig1}, the proposed framework consists of a multi-scale image feature extraction branch and a fine-grained distortion perception branch, starting from the input distorted image. First, SAM is employed to automatically segment semantically distinct regions.  Then, the MLLM generates semantic content, multi-dimensional distortion ratings and scores from the selected regions.  Subsequently, textual features are extracted using a text tokenizer and a text encoder. In the other branch, image features at multiple levels are extracted using the backbone and subsequently fused through the Multi-scale Hierarchical Fusion (MHF) module. Next, the Region-aware Semantic Attention (RSA) module incorporates region-level semantic guidance into the image features. Finally, region-level semantic content and multi-dimensional distortion information are aggregated to enable fine-grained assessment of the perceptual quality of the distorted image.
\subsection{Model Design}
\begin{figure}[t]
\centering
\includegraphics[width=0.9\columnwidth]{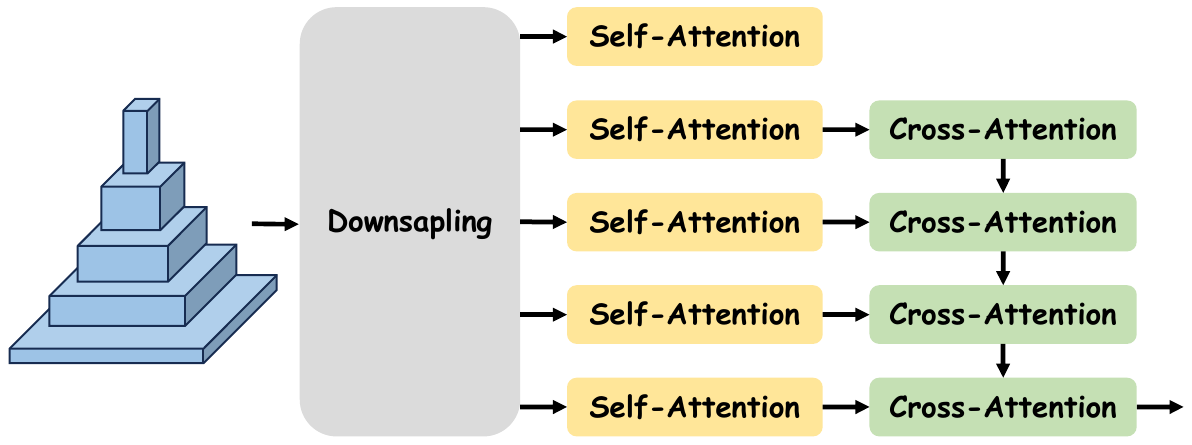} 
\caption{The architecture of MHF. The downsampling module downsamples the features from all levels to the resolution of the highest-level features.}
\label{MHF}
\end{figure}
\subsubsection{Multi-scale Hierarchical Fusion}
As illustrated in Figure \ref{MHF}, denote input image as $I\in{\mathbb{R}^{H\times W\times 3}}$. The backbone extracts multi-level features, where the feature of the $i$-th block is represented as $F_i\in{\mathbb{R}^{H_i\times W_i\times C_i}}$, where $H_i$, $W_i$ are height and width, $C_i$ is the channel dimension, $i\in\{1,2,...,n\}$. In general, low-level features are twice as large as their adjacent high-level features, and we have $H_i = H / 2^i$. Therefore, directly computing large matrices such as $F_1$ and $F_2$ is too expensive.  For simplicity and efficiency, we downsample $F_i$ to simplify it to the same shape as the highest-level feature $F_n$. We extract distortion-related features while suppressing redundant information using gated convolution \cite{yu2019free}. Accordingly, the downsampling operation is formulated as:
\begin{equation}
	D_i=\sigma(\text{Conv}(\text{Pool}(\sigma(\phi_i(F_i))\cdot(W_fF_i))))
\end{equation}
where $\sigma$ is the sigmoid activation function that constrains the mask value to the range of $(0, 1)$, $\phi_i$ represents a bottleneck convolution block, and $\text{Pool}(\cdot)$ denotes average pooling. The downsampling operation generates multi-level features $D_i \in \mathbb{R}^{H_n \times W_n \times C}$, where $C$ denotes the dimensionality of the reduced features. 

To further improve model efficiency, we adopt scaled dot-product attention to integrate multi-scale information and reduce computational complexity. Given a set of feature vectors consisting of a query ($Q$), a key ($K$), and a value ($V$), the attention mechanism computes the similarity between $Q$ and $K$ to generate a weighted sum of $V$.  This process can be formulated as:
\begin{equation}
	\text{Attn}(Q,K,V)=\text{softmax}(\frac{QK^T}{\sqrt{d_k}})V
    \label{func2}
\end{equation}
After downsampling, we obtain a set of multi-scale features, denoted as $\{D_1, D_2, \ldots, D_n\}\in{\mathbb{R}^{H_n\times W_n\times C}}$. Owing to the limited receptive field of low-level features, we first enhance $D_i$ using self-attention, as formulated below:
\begin{equation}
	D_i^{\prime}=\text{Attn}(D_i,D_i,D_i)+D_i
\end{equation}
where $D_i$ is projected onto $Q$, $K$, and $V$ through a simple linear projection.  Through self-attention, $D^\prime_i$ aggregates features from other positions to enhance $D_i$. Then, we apply cross-attention to integrate multi-scale features, as follows:
\begin{equation}
	D_i^{\prime \prime}=\text{Attn}(D_{i+1}^{\prime \prime},D_i^{\prime},D_i^{\prime})+D_{i+1}^{\prime \prime}
\end{equation}
where $i\in\{1,2,...,n-1\}$, and $D_n^{\prime \prime}=D_n^{\prime}$. The output of MHF, $D_1^{\prime\prime}$, is obtained by successively applying cross-attention across multiple levels.
\subsubsection{Fine-grained Distortion Perception}
To enable fine-grained distortion perception, the input image $I$ is segmented into semantic regions in a parallel branch, as shown in Figure \ref{fig1}, to preserve structural continuity while suppressing the influence of irrelevant regions. Semantic region segmentation is based on a powerful large-scale foundation model: SAM, which can accurately segment any object in any image without additional training. The segmentation mask generated by SAM can be formulated as:
\begin{equation}
	M=\text{Postprocess}(\text{SAM}(I))
\end{equation}
where $\text{Postprocess}(\cdot)$ ensures that the segmentation masks $M \in \mathbb{R}^{H \times W \times L}$ divide the image into $L$ non-overlapping regions. It ranks the masks produced by $\text{SAM}(\cdot)$ in descending order of their Predicted IoU Scores, where a higher score is assumed to correspond to more semantically meaningful regions. We retain the top $(L{-}1)$ masks with the highest scores. Pixels belonging to multiple masks are assigned to the one with the highest Predicted IoU Score, while those not covered by the top $(L{-}1)$ masks are designated as the background region, resulting in a total of $L$ semantic regions.

Subsequently, each segmented region is fed into the MLLM, and according to the prompt words obtains the content of this semantic region along with multidimensional distortion ratings and corresponding scores, where the distortion dimensions are $\{color, noise, artifact, blur, overall\}$. For each of the $L$ semantic regions, its content, multi-dimensional distortion ratings and scores are combined into a fine-grained description.  These $L$ descriptions are then fed into the Text Tokenizer and Text Encoder to obtain the text embedding matrix $E \in \mathbb{R}^{n \times d}$.
\subsubsection{Region-aware Semantic Attention}\label{RSA}
To integrate the segmented regions with semantic information, the semantic embeddings are used as query ($Q$) to attend to the segmented regions, which serve as $K$ and $V$, producing region-level semantically guided representations, as follows:
\begin{equation}
	G_i=\text{Attn}(E_i,M_i\cdot I,M_i\cdot I)+M_i\cdot I
\end{equation}
where $G_i\in{\mathbb{R}^{H\times W\times C_G}}$, and $M_i \cdot I$ denotes the $i$-th segmented semantic region, obtained by applying the mask $M_i$ to the input image $I$, where $i \in \{1, 2, ..., L\}$. Subsequently, RSA incorporates the region-level semantically guided representations into the image features generated by MHF by introducing a learnable attention bias.  The attention bias $B$ is computed as follows:
\begin{align}
    G_i^{\prime} &= \text{Reshape}(\text{Interp}(G_i)) \label{eq:reshape} \\
    B &= \lambda \sum_{i=1}^{n} G_i G_i^{T} \label{eq:covariance}
\end{align}
where Eq.~\eqref{eq:reshape} interpolates $G_i$ to the target spatial resolution $H^{\prime} \times W^{\prime}$ and reshapes the result to $G_i^{\prime} \in \mathbb{R}^{H^{\prime} W^{\prime} \times C_G}$. Equation~\eqref{eq:covariance} subsequently derives the learnable attention bias $B\in \mathbb{R}^{H^{\prime} W^{\prime} \times H^{\prime} W^{\prime}}$ from the correlations among pixels within each semantic region, where $\lambda$ denotes the regularization coefficient. Therefore, given $Q, K, V \in \mathbb{R}^{H_n \times W_n \times C}$, which are the query, key, and value from the input feature $D_1^{\prime\prime} \in \mathbb{R}^{H_n \times W_n \times C}$, the mechanism of RSA can be formulated as follows:
\begin{equation}
    R = \text{softmax}\left(B+\frac{QK^T}{\sqrt{d_k}}\right) V
\end{equation}
where $d_k$ is the vector dimension of $Q$, $K$, and $V$. It is worth noting that we control $H^\prime W^\prime = H_n = W_n$ to ensure spatial alignment. The region-level semantic guidance matrix $B$ dynamically adjusts the interaction weights between different semantic regions, enhancing the attention correlation of relevant regions while suppressing the interference of irrelevant regions, thereby improving the model’s perception of local details.
\subsubsection{Score Regression and Loss Function}
The final scores are obtained using the final features $R$ as follows:
\begin{equation}
    \hat{y} = \text{MLP}(\text{Pool}(\text{Self-Attn}(R)))
\end{equation}
where Self-Attn is a self-attention block, followed by average pooling. It is introduced to better aggregate features from all positions. $\text{MLP}(\cdot)$ denotes a multilayer perceptron. We use datasets labeled with Mean Opinion Score (MOS), and we first normalize the MOS to $[0, 1]$, then use MSE loss:
\begin{equation}
	L(\hat{y},y)=\frac{1}{N}\sum_{i=1}^{N}\Vert\hat{y}-y\Vert^2
\end{equation}
\begin{table*}[t]
\centering
\resizebox{2.10\columnwidth}{!}{
\begin{tabular}{c|c c c c c c c}
    \toprule
    Method & LIVE & CSIQ & TID2013 & KADID-10k & CLIVE & KonIQ-10k & SPAQ\\
    \midrule
    NIQE (SPL 2012) & 0.906/0.902 & 0.808/0.806 & 0.640/0.519 & 0.558/0.528 & 0.493/0.451 & 0.534/0.526 & 0.712/0.713\\
    BRISQUE (TIP 2012) & 0.944/0.929 & 0.748/0.812 & 0.571/0.626 & 0.567/0.528 & 0.629/0.629 & 0.685/0.681 & 0.817/0.809 \\
    \midrule
    HyperIQA (CVPR 2020) & 0.966/0.962 & 0.847/0.833 & 0.858/0.840 & 0.845/0.852 & 0.882/0.859 & 0.917/0.906 & 0.915/0.911\\
    MUSIQ (ICCV 2021) & 0.911/0.940 & 0.876/0.857 & 0.815/0.773 & 0.872/0.875 & 0.746/0.702 & 0.928/0.916 & 0.921/0.918 \\
    ManIQA (CVPRW 2022) & 0.978/0.976 & 0.954/0.949 & 0.938/0.931 & 0.941/0.930 & 0.867/0.840 & 0.849/0.834 & 0.893/0.872 \\
    CLIP-IQA+ (AAAI 2023) & 0.952/0.936 & 0.929/0.918 & 0.856/0.933 & 0.901/0.885 & 0.822/0.804 & 0.909/0.895 & 0.728/0.733\\
    Re-IQA (CVPR 2023) & 0.971/0.970 & 0.960/0.947 & 0.861/0.804 & 0.885/0.872 & 0.854/0.840 & 0.923/0.914 & \underline{0.925/0.918}\\
    PCIQA (TMM 2024) & 0.976/0.976 & 0.942/0.931 & 0.883/0.857 & 0.866/0.870 & 0.880/0.863 & 0.930/0.918 & 0.908/0.904\\
    LoDa (CVPR 2024) & 0.979/0.975 & 0.958/0.957 & 0.901/0.869 & 0.936/0.931 & 0.889/0.876 & 0.934/0.927 & 0.920/0.915\\
    QCN (CVPR 2024) & 0.980/0.974 & 0.967/0.960 & 0.892/0.875 & 0.924/0.919 & \underline{0.893/0.875} & 0.936/0.928 & 0.915/0.913\\
    Nguyen et al. (WACV 2025) & 0.972/0.969 & \underline{0.968/0.964} & 0.883/0.863 & 0.925/0.930 & 0.867/0.840 & 0.925/0.915 & 0.922/0.911\\
    VISGA (TCSVT 2025) & \underline{0.981/0.979} & 0.964/0.959 & 0.914/0.901 & 0.925/0.919 & 0.893/0.843 & 0.937/0.930 & 0.921/0.909\\
    \midrule
    RSFIQA (ResNet50) & \textbf{0.983/0.981} & \textbf{0.972/0.968} & \underline{0.957/0.946} & \underline{0.949/0.945} & 0.872/0.833 & \underline{0.937/0.932} & 0.923/0.916\\
    std & 0.002/0.002 & 0.003/0.002 & 0.010/0.012 & 0.011/0.013 & 0.014/0.012 & 0.003/0.003 & 0.012/0.011\\
    RSFIQA (Swin) & 0.980/0.978 & 0.959/0.960 & \textbf{0.959/0.951} & \textbf{0.954/0.953} & \textbf{0.900/0.897} & \textbf{0.940/0.934} & \textbf{0.929/0.923}\\
    std & 0.003/0.002 & 0.003/0.002 & 0.011/0.009 & 0.012/0.012 & 0.013/0.010 & 0.003/0.002 & 0.012/0.010\\
    \bottomrule
\end{tabular}}
\caption{Comparison of intra-dataset evaluation results based on PLCC/SRCC metrics. In each column, the \textbf{best} and \underline{second}-best results are highlighted in \textbf{bold} and \underline{underlined}, respectively.}
\label{tab1}
\end{table*}
\begin{table*}[t]
\centering
\resizebox{2.10\columnwidth}{!}{
\begin{tabular}{c|c c c c c c}
    \toprule
    Method & LIVE & CSIQ & TID2013 & KADID-10k & CLIVE & SPAQ\\
    \midrule
    NIQE (SPL 2012) & 0.793/0.677 & 0.718/0.628 & 0.651/0.568 & 0.468/0.405 & 0.642/0.609 & 0.679/0.664\\
    BRISQUE (TIP 2012) & 0.724/0.630 & 0.740/0.556 & 0.633/0.515 & 0.429/0.356 & 0.613/0.589 & 0.490/0.406\\
    \midrule
    HyperIQA (CVPR 2020) & 0.782/0.745 & 0.752/0.717 & 0.702/0.682 & 0.506/0.468 & 0.698/0.661 & 0.791/0.788\\
    MUSIQ (ICCV 2021) & 0.809/0.786 & 0.771/0.710 & 0.743/0.725 & 0.575/0.556 & 0.558/0.530 & 0.868/0.863\\
    ManIQA (CVPRW 2022) & 0.686/0.674 & 0.623/0.627 & 0.580/0.562 & 0.499/0.465 & 0.521/0.519 & 0.768/0.758\\
    CLIP-IQA+ (AAAI 2023) & 0.813/0.784 & 0.772/0.719 & 0.745/0.727 & 0.653/0.654 & 0.832/0.805 & 0.866/0.864\\
    LoDa (CVPR 2024) & 0.821/0.809 & 0.784/0.787 & 0.735/0.728 & 0.666/0.657 & 0.790/0.789 & 0.874/0.852\\
    Nguyen et al. (WACV 2025) & 0.823/0.821 & 0.774/0.765 & 0.743/0.735 & 0.672/0.660 & 0.793/0.781 & 0.875/0.863\\
    VISGA (TCSVT 2025) & 0.832/0.828 & 0.777/0.769 & 0.749/0.744 & 0.681/0.676 & 0.815/0.810 & 0.881/0.873\\
    \midrule
    Q-Align (ICML 2024) & \underline{0.839/0.831} & 0.785/0.737 & 0.757/0.746 & 0.674/0.684 & 0.853/0.860 & 0.886/0.887\\
    DeQA (CVPR 2025) & 0.835/0.830 & \underline{0.787/0.744} & 0.759/0.750 & \textbf{0.694/0.687} & \underline{0.858/0.855} & 0.884/0.886\\
    \midrule
    RSFIQA (ResNet50) & \textbf{0.844/0.836} & 0.786/0.740 & \underline{0.761/0.757} & 0.686/0.684 & 0.852/0.849 & \underline{0.889/0.887}\\
    RSFIQA (Swin) & 0.833/0.827 & \textbf{0.791/0.749} & \textbf{0.764/0.758} & \underline{0.689/0.685} & \textbf{0.861/0.859} & \textbf{0.892/0.889}\\
    \bottomrule
\end{tabular}}
\caption{Comparison of cross-dataset evaluation results based on PLCC/SRCC metrics.}
\label{tab2}
\end{table*}
\subsection{Prompt Format}
Tokens of segmented sub-images are denoted as $<$img$>$, sub-image content as $<$content$>$, distortion level as $<$level$>$, distortion score as $<$score$>$, and distortion type as $<$type$>$. The prompt format is defined as follows:

\textit{\#User: }$<$img$>$ \textit{Please describe the content of the highlighted area in the mask image based on the original image context.}

\textit{\#Assistant: }$<$content$>$.

\textit{\#User: }$<$img$>$ \textit{From the }$<$type$>$ \textit{dimension, please provide the quality level for highlighted area in the mask image.}

\textit{\#Assistant: From the} $<$type$>$ \textit{dimension, the image quality level is} $<$level$>$.

\textit{\#User: }$<$img$>$ \textit{From the} $<$type$>$ \textit{dimension, please provide the quality score for highlighted area in the mask image.}

\textit{\#Assistant: From the} $<$type$>$ \textit{dimension, the image quality score is} $<$score$>$.
\section{Experiments}
\subsection{Implementation Details}
\subsubsection{Datasets}
For a comprehensive evaluation, we include a wide range of IQA datasets: LIVE \cite{sheikh2006statistical}, CSIQ \cite{larson2010most}, TID2013 \cite{ponomarenko2013color}, KADID-10k \cite{lin2019kadid}, CLIVE \cite{ghadiyaram2015massive}, KonIQ-10k \cite{hosu2020koniq} and SPAQ \cite{fang2020perceptual}. The first four are synthetic distortion datasets, whereas the latter three contain authentic distortions. We use the official train/val/test splits if available, otherwise, we randomly split it 10 times and report the mean and variance.
\begin{figure*}[t]
\centering
\includegraphics[width=2.10\columnwidth]{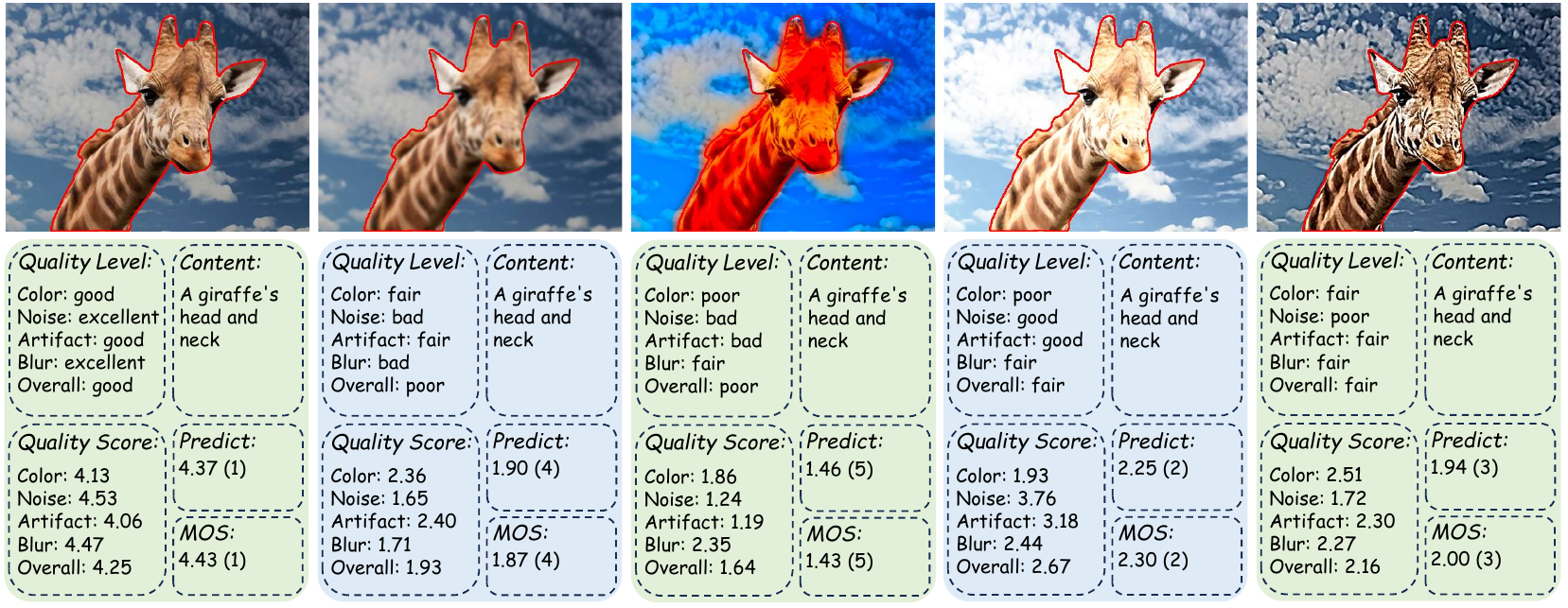} 
\caption{Example images from KADID-10k.  The region outlined in red in the image is one of the semantic regions segmented by SAM.  Below the image are the corresponding outputs of MLLM, including the content of the segmented region, multi-dimensional distortion ratings and scores.  MOS indicates the ground-truth human rating, and the number in parentheses represents the ranking of the distorted image considered in this example.}
\label{fig4}
\end{figure*}
\begin{table*}[t]
\centering
\resizebox{2.10\columnwidth}{!}{
\begin{tabular}{c c c|c c c c c c c}
    \toprule
    MHF & MLLM & RSA & LIVE & CSIQ & TID2013 & KADID-10k & CLIVE & KonIQ-10k & SPAQ\\
    \midrule
     &  &  & 0.956/0.957 & 0.934/0.931 & 0.930/0.924 & 0.927/0.922 & 0.874/0.868 & 0.909/0.902 & 0.899/0.892\\
    \usym{2714} &  &  & 0.961/0.957 & 0.938/0.936 & 0.935/0.930 & 0.934/0.931 & 0.873/0.864 & 0.912/0.910 & 0.905/0.901\\
    \usym{2714} & \usym{2714} &  & \underline{0.971/0.967} & 0.949/0.946 & \underline{0.951/0.943} & \underline{0.945/0.950} & \underline{0.890/0.883} & 0.924/0.918 & \underline{0.916/0.910}\\
    \usym{2714} &  & \usym{2714} & 0.969/0.966 & \underline{0.950/0.949} & 0.945/0.941 & 0.944/0.942 & 0.883/0.879 & \underline{0.926/0.920} & 0.915/0.912\\
    \usym{2714} & \usym{2714} & \usym{2714} & \textbf{0.980/0.978} & \textbf{0.959/0.960} & \textbf{0.959/0.951} & \textbf{0.954/0.953} & \textbf{0.900/0.897} & \textbf{0.940/0.934} & \textbf{0.929/0.923}\\
    \bottomrule
\end{tabular}}
\caption{Ablation study on MHF, MLLM, and RSA based on PLCC/SRCC metrics. }
\label{tab3}
\end{table*}
\subsubsection{Metrics}
we use the Pearson Linear Correlation Coefficient (PLCC) and Spearman Rank-order
Correlation Coefficient (SRCC) as metrics to evaluate the score regression performance. PLCC measures the linear correlation between predicted
scores $\hat{y}$ and ground truth labels $y$, while SRCC assesses
rank correlation. The calculation formula of PLCC is as follows:
\begin{equation}
\mathrm{PLCC} = \frac{
    \sum_{i=1}^{n} (y_i - \bar{y})(\hat{y}_i - \bar{\hat{y}})
}{
    \sqrt{\sum_{i=1}^{n} (y_i - \bar{y})^2} \cdot \sqrt{\sum_{i=1}^{n} (\hat{y}_i - \bar{\hat{y}})^2}
}
\label{eq:plcc}
\end{equation}
where $\bar{y}$ and $\bar{\hat{y}}$ represent the mean values of $y$ and $\hat{y}$, respectively.
\subsubsection{Training Details}
For the image encoder, we adopt a pretrained ResNet \cite{he2016deep} or Swin Transformer \cite{liu2021swin} as the backbone network, and resize all input images to $224 \times 224$. For the text encoder, we use a modified transformer \cite{radford2019language} pretrained within CLIP \cite{radford2021learning}. The MLLM employed in RSFIQA is Qwen2.5-VL \cite{bai2025qwen2}. We set $L$ to 5, and we use the AdamW optimizer with a weight decay of $10^{-5}$ for all experiments. The initial learning rate (lr) is set to $3 \times 10^{-5}$. We use a cosine annealing scheduler with $T_{max}=50$, $\eta_{min}=0$, $\eta_{max}=lr$. The total number of training epochs was set to 200, and an early stopping strategy based on validation performance was employed to reduce unnecessary training time. Our model is implemented using PyTorch and trained on an NVIDIA GeForce RTX 3060 GPU.
\begin{table*}[t]
\centering
\resizebox{2.10\columnwidth}{!}{
\begin{tabular}{c c c|c c c c c c c}
    \toprule
    Content & Level & Score & LIVE & CSIQ & TID2013 & KADID-10k & CLIVE & KonIQ-10k & SPAQ\\
    \midrule
    &  &  & 0.970/0.965 & 0.949/0.947 & 0.948/0.944 & 0.944/0.942 & 0.881/0.877 & 0.925/0.927 & 0.913/0.909\\
    \usym{2714} &  &  & 0.973/0.968 & 0.950/0.951 & 0.948/0.944 & 0.947/0.945 & 0.887/0.880 & 0.930/0.927 & 0.919/0.916\\
    \usym{2714} & \usym{2714} &  & 0.975/0.972 & \textbf{0.959/0.955} & 0.953/0.950 & \underline{0.952/0.950} & \underline{0.892/0.889} & 0.936/0.931 & 0.922/0.917\\
    \usym{2714} &  & \usym{2714} & \underline{0.978/0.974} & 0.957/0.956 & \underline{0.955/0.949} & 0.951/0.953 & 0.891/0.887 & \textbf{0.940/0.932} & \underline{0.926/0.925}\\
    \usym{2714} & \usym{2714} & \usym{2714} & \textbf{0.980/0.978} & \textbf{0.959/0.960} & \textbf{0.959/0.951} & \textbf{0.954/0.953} & \textbf{0.900/0.897} & \textbf{0.940/0.934} & \textbf{0.929/0.923}\\
    \bottomrule
\end{tabular}}
\caption{Ablation study on prompt format based on PLCC/SRCC metrics.}
\label{tab4}
\end{table*}
\begin{table*}[t]
\centering
\resizebox{2.10\columnwidth}{!}{
    \begin{tabular}{c c c c c|c c c c c c}
        \toprule
        Color & Noise & Artifact & Blur & Overall & LIVE & TID2013 & KADID-10k & CLIVE & KonIQ-10k & SPAQ\\
        \midrule
         & \usym{2714} & \usym{2714} & \usym{2714} & \usym{2714} & 0.974/0.971 & 0.954/0.949 & 0.946/0.943 & 0.893/0.891 & 0.936/0.930 & 0.924/0.916\\
        \usym{2714} &  & \usym{2714} & \usym{2714} & \usym{2714} & \underline{0.977/0.975} & 0.955/0.958 & 0.948/0.945 & 0.894/0.891 & 0.936/0.933 & 0.923/0.917\\
        \usym{2714} & \usym{2714} &  & \usym{2714} & \usym{2714} & 0.972/0.968 & 0.954/0.943 & \underline{0.950/0.948} & 0.893/0.889 & 0.933/0.927 & 0.924/0.914\\
        \usym{2714} & \usym{2714} & \usym{2714} &  & \usym{2714} & 0.974/0.973 & \underline{0.957/0.950} & 0.945/0.941 & 0.896/0.893 & 0.935/0.932 & 0.924/0.918\\
        \usym{2714} & \usym{2714} & \usym{2714} & \usym{2714} &  & 0.976/0.972 & 0.954/0.947 & 0.949/0.946 & \textbf{0.900/0.895} & \underline{0.938/0.931} & \underline{0.925/0.920}\\
        \usym{2714} & \usym{2714} & \usym{2714} & \usym{2714} & \usym{2714} & \textbf{0.980/0.978} & \textbf{0.959/0.951} & \textbf{0.954/0.953} & \textbf{0.900/0.897} & \textbf{0.940/0.934} & \textbf{0.929/0.923}\\
        \midrule
        \bottomrule
    \end{tabular}}
\caption{Ablation study on multiple distortion dimensions based on PLCC/SRCC metrics.}
\label{dimensions}
\end{table*}
\begin{table}[t]
\centering
\resizebox{1.00\columnwidth}{!}{
\begin{tabular}{c|c c c c c}
    \toprule
    $L$ & 3 & 4 & 5 & 6\\
    \midrule
    LIVE & 0.973/0.970 & 0.976/0.977 & \textbf{0.980/0.978} & \underline{0.978/0.976}\\
    KADID-10k & 0.947/0.944 & 0.952/0.950 & \underline{0.954/0.953} & \textbf{0.956/0.953}\\
    CLIVE & 0.892/0.889 & \underline{0.899/0.894} & \textbf{0.900/0.897} & 0.898/0.896\\
    KonIQ-10k & 0.934/0.925 & 0.937/0.930 & \textbf{0.940/0.934} & \underline{0.939/0.932}\\
    \midrule
    Inference Time (s) & \textbf{0.092} & \underline{0.103} & 0.136 & 0.195\\
    \bottomrule
\end{tabular}}
\caption{Ablation study on the number of segmented semantic regions $L$ based on PLCC/SRCC metrics.}
\label{tab5}
\end{table}
\subsection{Results of Quality Assessment}
\subsubsection{Intra-dataset Results}
We first conduct intra-dataset experiments, and the corresponding experimental results are shown in Table~\ref{tab1}. Our method achieves superior performance on both synthetic and authentic distortion datasets compared to all baseline methods. For instance, compared with VISGA \cite{shi2025visual}, both PLCC and SRCC achieve an average improvement of 1.0\% across the two types of datasets, demonstrating the effectiveness and robustness of our approach under diverse distortion conditions.
\subsubsection{Cross-dataset experiments}
In addition, we train our model on KonIQ-10k dataset and evaluate its performance on several out-of-distribution datasets, following the experimental protocol in \cite{wu2023q}. The results are shown in Table~\ref{tab2}. Models based on MLLMs generally outperform traditional IQA methods. Our RSFIQA achieves consistently superior or competitive performance across all out-of-distribution datasets, further demonstrating its robustness and strong generalization capability in capturing fine-grained perceptual quality variations under diverse and unseen conditions.
\subsubsection{Qualitative Results}
As shown in Figure~\ref{fig4}, across different types of distortions applied to the same reference image, our method consistently produces quality scores that closely align with human evaluations. The quality ranking of these five distorted images matches human perception, highlighting the reliability and fine-grained distortion sensitivity of the proposed quality assessment approach.
\subsection{Ablation Study}
\subsubsection{Ablation Study on Different Components}In table \ref{tab3}, we conduct ablation studies on the components proposed in RSFIQA. The baseline is a simple regression network that extracts multi-scale features using a Swin Transformer, and each proposed component is added sequentially. We conduct ablation studies to evaluate the contribution of the three core components of RSFIQA: (1) Multi-scale Hierarchical Fusion (MHF), (2) Multi-modal Large Language Model (MLLM), and (3) Region-aware Semantic Attention (RSA). We observe that all three components contribute positively to the performance. Specifically, MHF yields a modest improvement over the baseline, while MLLM and RSA offer more substantial gains, highlighting the effectiveness of region-level semantic guidance.
\subsubsection{Ablation Study on Prompts}
We conducte ablation studies on three different types of prompts: (1) Content, (2) Quality Level, and (3) Quality Score, as shown in Table \ref{tab4}. For the first row, we directly set the MLLM output to “Answer.”. The results indicate that the combination of all three types of prompts achieves the best performance, as it integrates both the content within the segmented regions and fine-grained descriptions of multi-dimensional distortion perception.
\subsubsection{Ablation study on multiple distortion dimensions.}
We demonstrate the effectiveness of multiple distortion dimensions, as shown in Table \ref{dimensions}. It can be observed that applying all distortion dimensions achieves the best performance. In addition, interestingly, after removing the overall dimension, the model performs the second-best, further illustrating the importance of fine-grained perception based on multiple distortion dimensions.
\subsubsection{Ablation Study on the Number of Segmented Regions}
We evaluate the effect of the parameter $L$, which controls the number of semantic regions segmented by SAM, as shown in Table~\ref{tab5}. Increasing $L$ allows the model to capture finer semantic details by dividing the image into more regions, potentially improving the granularity of quality assessment. Our results indicate that setting $L = 5$ achieves PLCC and SRCC metrics comparable to those obtained with $L = 6$, while significantly reducing inference time. Thus, we set $L$ to $5$ for a better trade-off between accuracy and efficiency.
\section{Conclusion}
In this paper, we propose RSFIQA, a novel region-level semantic-guided fine-grained no-reference image quality assessment method. By leveraging the strengths of SAM for semantic segmentation and MLLM for multi-dimensional distortion analysis, our approach achieves precise and comprehensive quality assessment across distinct semantic regions. To effectively fuse semantic guidance with image features for quality regression, we introduce a Region-level Semantic Aggregation (RSA) module. Extensive experiments on widely-used IQA benchmarks demonstrate that RSFIQA consistently outperforms existing state-of-the-art methods, validating its effectiveness and robustness. Future work will focus on extending this framework to more challenging scenarios such as video quality assessment, aiming to address temporal consistency and dynamic distortion complexities.



\bibliography{aaai2026}
\end{document}